\documentclass[dvipsnames, format=sigconf, anonymous=false, review=false, nonacm=true]{acmart}

\AtBeginDocument{%
  \providecommand\BibTeX{{%
    \normalfont B\kern-0.5em{\scshape i\kern-0.25em b}\kern-0.8em\TeX}}}

\setcopyright{acmcopyright}
\copyrightyear{2018}
\acmYear{2018}
\acmDOI{10.1145/1122445.1122456}

\acmConference[Woodstock '18]{Woodstock '18: ACM Symposium on Neural
  Gaze Detection}{June 03--05, 2018}{Woodstock, NY}
\acmBooktitle{Woodstock '18: ACM Symposium on Neural Gaze Detection,
  June 03--05, 2018, Woodstock, NY}
\acmPrice{15.00}
\acmISBN{978-1-4503-XXXX-X/18/06}

\newcommand{\replications}[0]{5}
\newcommand{\uqdlong}[0]{Uncertain Quality-Diversity}
\newcommand{\uqd}[0]{UQD}
\newcommand{\code}[0]{URL
\url{https://github.com/adaptive-intelligent-robotics/QDax}
}
\newcommand{\location}[0]{uncertainty location}
\newcommand{\Location}[0]{Uncertainty location}

\renewcommand{\vec}[1]{{\bm{{#1}}}} 
\newcommand{\mat}[1]{{\bm{{#1}}}} 

\usepackage[ruled, lined, linesnumbered, commentsnumbered, noend]{algorithm2e}
\usepackage{xcolor}
\usepackage{soul}  

\usepackage{enumitem}
\usepackage{microtype}
\usepackage[skip=6pt]{caption}
\usepackage{makecell}
\usepackage{bm}
\usepackage{multirow}

\begin{document}


\title{
Benchmark tasks for Quality-Diversity applied to Uncertain domains
}

\author{Manon Flageat}
\email{manon.flageat18@ic.ac.uk}
\affiliation{%
  \institution{Imperial College London}
  \city{London}
  \country{U.K.}
}

\author{Luca Grillotti}
\email{luca.grillotti16@imperial.ac.uk}
\affiliation{%
  \institution{Imperial College London}
  \city{London}
  \country{U.K.}
}

\author{Antoine Cully}
\email{a.cully@imperial.ac.uk}
\affiliation{%
  \institution{Imperial College London}
  \city{London}
  \country{U.K.}
}


\begin{abstract}

While standard approaches to optimisation focus on producing a single high-performing solution, Quality-Diversity (QD) algorithms allow large diverse collections of such solutions to be found. If QD has proven promising across a large variety of domains, it still struggles when faced with uncertain domains, where quantification of performance and diversity are non-deterministic. 
Previous work in Uncertain Quality-Diversity (UQD) has proposed methods and metrics designed for such uncertain domains. 
In this paper, we propose a first set of benchmark tasks to analyse and estimate the performance of UQD algorithms. 
We identify the key uncertainty properties to easily define UQD benchmark tasks: the \location{}, the type of distribution and its parameters.
By varying the nature of those key UQD components, we introduce a set of $8$ easy-to-implement and lightweight tasks, split into $3$ main categories. 
All our tasks build on the Redundant Arm: a common QD environment that is lightweight and easily replicable.
Each one of these tasks highlights one specific limitation that arises when considering UQD domains.
With this first benchmark, we hope to facilitate later advances in UQD.

\end{abstract}


\ccsdesc[500]{Computational Methodologies ~Evolutionary Robotics}
\keywords{Quality-Diversity optimisation, Uncertain domains, MAP-Elites, Behavioural diversity.}


\maketitle

\section{Introduction}

Quality-Diversity (QD) optimisation \cite{framework, book_chapter} has proven in recent years to be a promising approach across multiple application domains \cite{nature, design, video_games_matt}.
QD aims to find a collection $\mathcal{A}$ of diverse and high-performing solutions to an optimisation problem. It does so by storing in $\mathcal{A}$ solutions $i$ with high-fitness $f_i$ that induce novel behaviours, quantified using behaviour descriptors $d_i$. In other words, QD aims to optimise:

\begin{align} \label{eq:qd_obj}
\begin{split}
    \max_{\mathcal{A}} \left( \textrm{QDScore} (\mathcal{A}) \right) = \max_{\mathcal{A}} \sum_{i \in \mathcal{A}}{f_i}\\
    \quad \text{w.r.t} \quad \forall i \in \mathcal{A}, d_i\in \textrm{cell}_i
\end{split}
\end{align}

\begin{figure}[t!]
\centering
\includegraphics[width = 0.85\hsize]{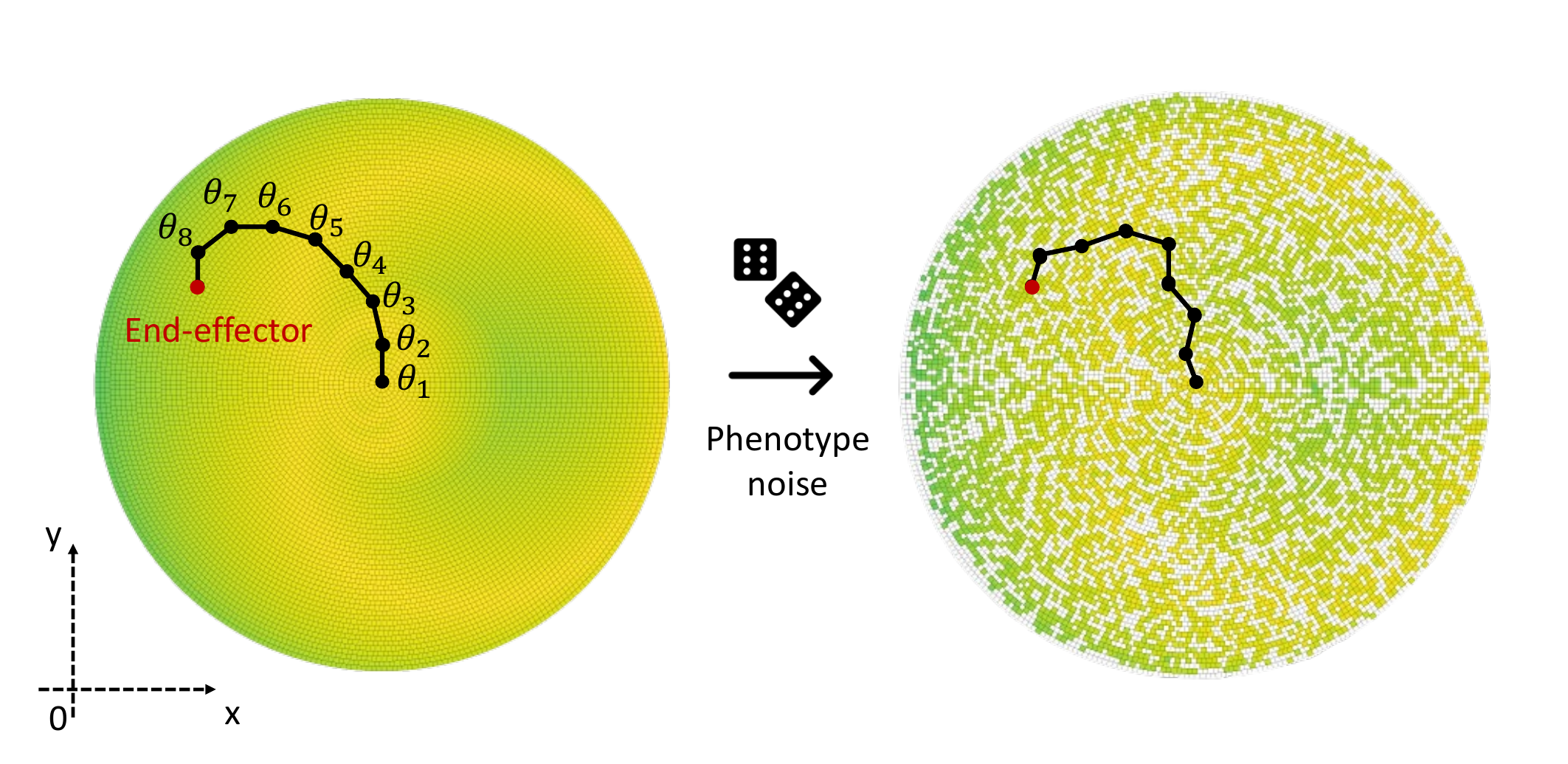}
\caption{
Illustration of the effect of uncertainty in the Redundant Arm QD task; in this case, the uncertainty is applied to the parameters of the arm.
The descriptor is given as the end-effector $(x, y)$ position and the fitness as the variance of the joint angles. The background figures the corresponding QD Illusory archive (left) and Corrected archive (right) after multiple reevaluations of each solution. 
}
\vspace{-4mm}
\label{fig:arm_noise}
\end{figure}

Despite their encouraging results, multiple works have high-lightened the limitations of QD algorithms when considering uncertain domains \cite{adaptive, deepgrid, uqd, aria}. 
A QD domain is seen as uncertain if a solution $i$ can get different fitness $f_i$ and descriptor $d_i$ values from one evaluation to another. In other words, in Uncertain Quality-Diversity (UQD) \cite{uqd} fitness and descriptor associated with one solution are no more fixed values but distributions: $\tilde{f_i} \sim \mathcal{D}_{f_i}$, $\tilde{d_i} \sim \mathcal{D}_{d_i}$.
When applied to such domains, usual QD algorithms such as MAP-Elites \cite{map_elites} tend to keep solutions that have been lucky during evaluation and get outlier fitness or descriptor values. This leads to a loss in the overall performance of their final collection.
Contrary to standard QD approaches, UQD algorithms take into consideration the distributional nature of their variables, and adapt the problem formulation accordingly \cite{uqd}:
\begin{equation} \label{eq:qd_uncertain_obj}
\begin{split}
    \max_{\mathcal{A}} \left( \textrm{QDScore} (\mathcal{A}) \right) = \max_{\mathcal{A}} \sum_{i \in \mathcal{A}}{\mathbb{E}_{\tilde{f_i} \sim \mathcal{D}_{f_i}} \left[ \tilde{f_i} \right]} \\
    \text{w.r.t} \quad \forall i \in \mathcal{A}, \mathbb{E}_{\tilde{d_i} \sim \mathcal{D}_{d_i}} \left[ \tilde{d_i} \right] \in \textrm{cell}_i
\end{split}
\end{equation}

\begin{table*}
\centering
\small
\caption{Performance-estimation Redundant Arm benchmark tasks.}
\vspace{-1mm}
\begin{tabular}{ l | l | l | l }

 & \textsc{\Location{}} & \textsc{Distribution} & \textsc{Parameters} \\ 
 \midrule 
 \addlinespace
 
 \textsc{Gaussian fitness noise} & \multirow{2}*{\makecell{Fitness after evaluation: $\begin{cases} \tilde{f} = f + \epsilon \\ \tilde{d} = d \end{cases}$}} & $\epsilon \sim \mathcal{N}(0, \sigma^2)$ & $\sigma > 0$ \\

 \textsc{Bimodal fitness noise} & & $\epsilon \sim \alpha \mathcal{N}(0, \sigma_1^2) + (1-\alpha) \mathcal{N}(-1, \sigma_2^2)$ & $\sigma_1, \sigma_2 >0 \quad\alpha \in (0, 1)$\\

\addlinespace
 \midrule 
\addlinespace

 \textsc{Small-Gaussian descriptor noise} & \multirow{3}*{\makecell{Descriptor after evaluation: $\begin{cases} \tilde{f} = f \\ \tilde{d} = d + \epsilon \end{cases}$}} & $\epsilon \sim \mathcal{N}(\vec 0, \sigma^2 \mat I)$ & $\sigma \in (0, 0.02)$ \\

 \textsc{Large-Gaussian descriptor noise} & & $\epsilon \sim \mathcal{N}(\vec 0, \sigma^2 \mat I)$ & $\sigma > 0.1$ \\

 \textsc{Bimodal descriptor noise} & & $\epsilon \sim \alpha \mathcal{N}(\vec 0, \sigma_1^2 \mat I) + (1-\alpha) \mathcal{N}(\left[\begin{smallmatrix} 1 \\ 1 \end{smallmatrix}\right], \sigma_2^2 \mat I)$ & $\sigma_1, \sigma_2 >0 \quad\alpha \in (0, 1)$ \\
 
\end{tabular}
\label{tab:performance_task}
\vspace{-2mm}
\end{table*}

\begin{table*}
\centering
\small
\caption{Reproducibility-maximisation Redundant Arm benchmark tasks.}
\vspace{-1mm}
\begin{tabular}{ l | l | l | l }

 & \textsc{\Location{}} & \textsc{Distribution} & \textsc{Parameters} \\ 
 \midrule 
 \addlinespace

 \makecell[l]{\textsc{Gaussian descriptor noise} \\ \textsc{with $2$ variance values}} & \multirow{2}[5]*{Descriptor after evaluation: $\begin{cases} \tilde{f} = 0 \\ \tilde{d} = d + \epsilon \end{cases}$} 
 & $\begin{cases}\epsilon \sim \mathcal{N}(\vec 0, \sigma_1^2 \mat I) \mbox{ if } sign(\prod_j{\theta^j}) \geq 0 \\ \epsilon \sim \mathcal{N}(\vec 0, \sigma_2^2 \mat I) \mbox{ otherwise} \end{cases}$ 
 & $\sigma_2 \gg \sigma_1 > 0$ \\

\addlinespace

\makecell[l]{\textsc{Gaussian descriptor noise} \\ \textsc{with continuous variance values}} & & $\epsilon \sim \mathcal{N}(\vec 0, (\eta * \mathbb{V}(\theta))^2 \mat I)$ & $\eta > 0$ \\
 
\end{tabular}
\label{tab:reproducibility_task}
\vspace{-2mm}
\end{table*}

\begin{table*}
\centering
\caption{Realistic Redundant Arm benchmark tasks.}
\vspace{-1mm}
\small
\begin{tabular}{ c | c | c | c }

 & \textsc{\Location{}} & \textsc{Distribution} & \textsc{Parameters} \\ 
 \midrule  

 \textsc{Gaussian Phenotype $J$ noise} 
 & 
 Dimension $J$ of the phenotype: $\tilde{\theta} =\begin{pmatrix} \theta^1 & \cdots & \theta^{J-1} &\theta^{J} + \epsilon & \theta^{J+1} & \cdots & \theta^N \end{pmatrix}$ & $\epsilon \sim \mathcal{N}\left(0, \sigma^2 \right)$ & $\sigma > 0$ \\
 
\end{tabular}
\label{tab:realistic_task}
\vspace{-2mm}
\end{table*}

Previous work in \uqd{} \cite{adaptive, uqd, deepgrid, aria} have proposed solutions tailored for these domains, as well as new metrics that better estimate the performance of QD algorithms when applied to \uqd{} domains. 
In this paper, we aim to propose a set of benchmark tasks to help future developments of \uqd{} algorithms. 
As first contribution, we propose a more systematic way to define \uqd{} benchmark tasks for a given environment, as the ensemble of (1) an \location{}, (2) an uncertainty distribution type and (3) its corresponding parameters. These properties also constitute initial tools to analyse inherent uncertainty in more realistic \uqd{} tasks, allowing to better select tailored approaches. 
As second contribution, we define, based on this definition, $8$ new tasks, split into $3$ main categories. These tasks aim to highlight some of the main limitations that arise in \uqd{} settings. 
All our tasks are based on the Redundant Arm task, a simple-to-implement and lightweight QD task that has been widely studied across the existing literature \cite{nature, benchmark}. 
To facilitate the use of our benchmark, we make our code available at \code{}.

\section{Properties of \uqd{} tasks}

One can define multiple \uqd{} benchmark tasks for a given environment by varying the properties of the uncertainty induced in the environment. 
%
We identified 3 main uncertainty properties: the \location{}, the type of distribution, and the parameters of this distribution.
In this section, we analyse those properties and explain how they can be used to define UQD benchmark tasks. 
They can also be used as tools to analyse more realistic \uqd{} tasks.

\subsection{\Location{}}

As explained above, in \uqd{} tasks, fitnesses and descriptors are no more fixed values but distributions. These distributions can arise from uncertainty at different levels within the environment. The clear definition of this \location{} is critical to define benchmark tasks and study \uqd{} tasks. 
Previous work in uncertain evolutionary algorithms proposed a classification that separates prior noise, applied to the phenotype before the fitness evaluation, and posterior noise, applied to the fitness after the evaluation~\cite{ea_uncertain_3}. 
In the case of \uqd{}, we propose to decompose further the prior noise category to encompass dynamic and applied problems as considered in QD. 
In simple QD environments such as Rastrigin or the Redundant Arm, one could consider adding (1) noise on the fitness and descriptor reading, or (2) noise on the phenotype (i.e. the parameters in Rastrigin and the joint angles in the Arm), which corresponds directly to the prior and posterior noise categories.
However, in more complex settings, such as robot control environments~\cite{nature}, one could consider adding (1) noise on the fitness and descriptor reading, (2) noise on the phenotype, i.e. the controller parameters, (3) noise on the initial position of the robot, and (4) stochasticity in the dynamics of the environment. 
Similar locations could be considered in closed-loop QD-RL environments \cite{benchmark, instadeep_benchmark}. 


\subsection{Distribution Type}

Another key property of \uqd{} tasks is the distribution of the noise or stochasticity. 
Most previous work in \uqd{} used Gaussian or Uniform distributions. 
However, more complex distributions such as skewed, multi-modal, or non-continuous distributions can lead to interesting properties as will be discussed later in this paper.
Depending on the \location{} within the environment, the structure of this distribution will then be shaped and modified to give the final fitness and descriptor distributions, leading to complex uncertainty structures.

\subsection{Distribution Parameters}

Varying the parameters of a distribution may also have a significant impact on the performance of UQD algorithms.
For example, gaussian descriptor distributions with a very low variance may have a negligible impact on an algorithm's results; indeed, if an individual's descriptor has a low variance, then most re-evaluations of that descriptor will land in the same behaviour niche (i.e. MAP-Elites cell).
Conversely, a really large fitness variance can make the comparison between two fitness values almost non-meaningful and thus prevent convergence. 

\section{Redundant Arm \uqd{} benchmark}

\subsection{Redundant Arm} \label{sec:arm}

\begin{figure}[t!]
\centering
\includegraphics[width = \hsize]{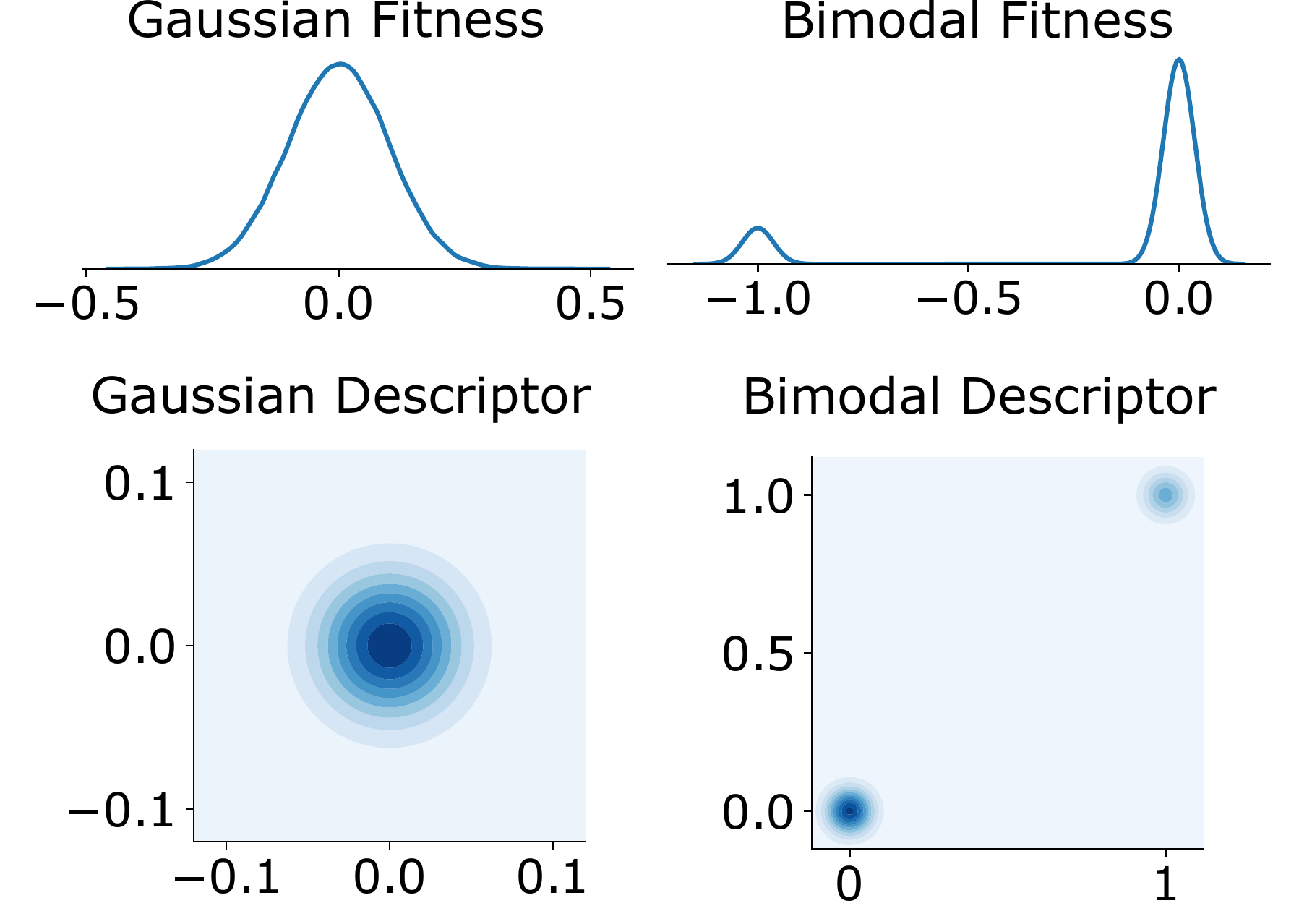}
\caption{
Distributions in the Performance-estimation tasks.
}
\vspace{-4mm}
\label{fig:arm_distributions}
\end{figure}

The Redundant Arm \cite{nature, framework, book_chapter}, illustrated in Figure~\ref{fig:arm_noise}, is one of the most commonly used QD benchmark task. It aims to learn how to achieve every reachable point with a planar $N$-DoF robotic Arm:
\begin{itemize}[leftmargin=*]
    \item \textbf{Controller: } real-valued angles of the $N$-DoF: $\theta_i=(\theta_i^j)_{1\leq j \leq N}$.
    \item \textbf{Descriptor: } defined as the $(x,y)$ position of the end-effector, after applying the control values in the joints.
    \item \textbf{Fitness: } in this paper, we consider two versions of this task:
    \begin{enumerate}[leftmargin=*]
        \item fitness $0$ for all individuals.
        \item fitness as negative variance of the $N$ joint angles $(\theta_i^j)_{1\leq j \leq N}$: 
$f_i=-\mathbb{V}(\theta_i) = -\frac{1}{N}\sum_{j=1}^{N}{(\theta_i^j-\overline{\theta_i})^2}$.
    \end{enumerate}
\end{itemize}

\subsection{\uqd{} benchmark tasks}

In the following, we propose a set of benchmark \uqd{} tasks in the Redundant Arm environment, chosen to illustrate some of the common challenges in \uqd{}. As highlighted by \citet{uqd}, \uqd{} approaches usually tackle two main problems: 
\begin{itemize}[leftmargin=*]
    \item \textbf{Performance estimation:} correctly estimating the expected fitness $\mbox{E}_{f_i \sim \mathcal{D}_f} \left[ f_i \right]$ and expected descriptor  $\mbox{E}_{d_i \sim \mathcal{D}_d} \left[ d_i \right]$ of solutions from their evaluations.
    \item \textbf{Reproducibility maximisation:} encouraging solutions that are more likely to get similar fitness and descriptor values from one evaluation to another. This corresponds to maintaining solutions with lower variance of fitness $\mbox{V}_{f_i \sim \mathcal{D}_f} \left[ f_i \right]$ and variance of each descriptor dimension $\mbox{V}_{d^j_i \sim \mathcal{D}_{d^j}} \left[ d^j_i \right]$. This issue only arises in tasks where solutions can have different variances.
\end{itemize}

In Section~\ref{sec:estimation}, we present $5$ benchmark tasks designed to assess the Performance estimation ability of \uqd{} algorithms; and in Section~\ref{sec:reproducibility}, we introduce $2$ other tasks to assess their ability to select reproducible solutions. We also propose in Section \ref{sec:realistic} a more realistic task that combines both problems. 
For each section, we give some preliminary results obtained by MAP-Elites and MAP-Elites-sampling (with $30$ samples) with $100\times100$ grid. These approaches are used as baselines in existing \uqd{} works \cite{deepgrid, adaptive, uqd, aria}.

\section{Performance Estimation tasks} \label{sec:estimation}

In this section, we propose benchmark tasks that assess the ability of \uqd{} approaches to estimate performance. 
We consider two cases: tasks where only the fitness is a distribution and tasks where only the descriptor is a distribution. One can easily combine these cases to get complex tasks with noise on both.
These tasks all build on the Redundant Arm with non-zero fitness from Section~\ref{sec:arm}. We introduce each task below, give corresponding equations in Table.~\ref{tab:performance_task} and Figure~\ref{fig:arm_distributions}, and some results in Figures~\ref{fig:perf_results} and~\ref{fig:performance_estimation_archives}. 

\begin{figure}[t]
\centering
\includegraphics[width = \hsize]{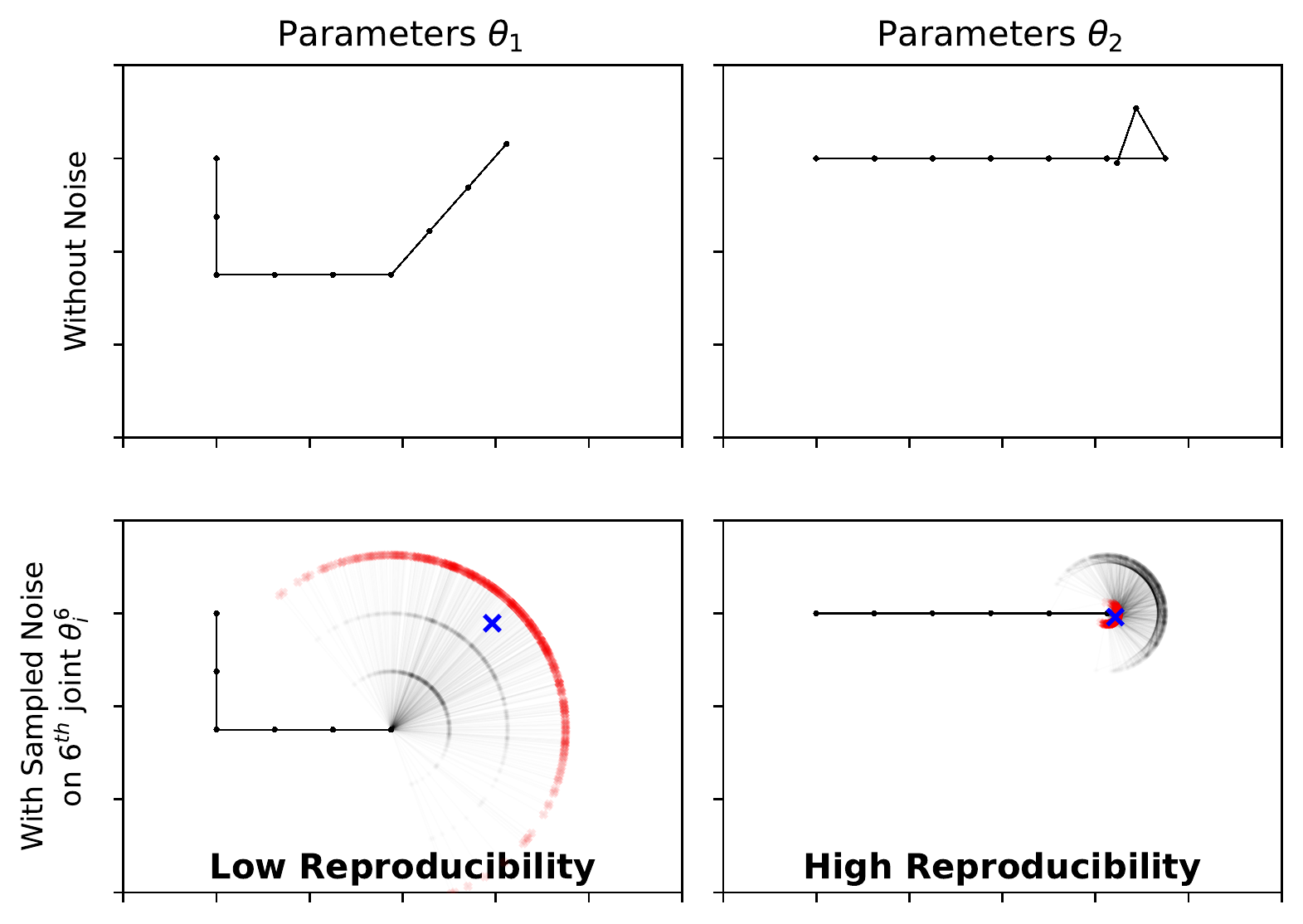}
\caption{
Illustration of the effect of noise applied on the $6^{\text{th}}$ joint of 2 distinct arm configurations $\theta_1$ (left) and $\theta_2$ (right).
The red crosses indicate the sampled descriptors, and the blue cross their average.
The descriptor reproducibility achieved by $\theta_2$ is way higher than the one achieved by $\theta_1$.
}
\vspace{-3mm}
\label{fig:arm_noise_on_6th_joint}
\end{figure}

\subsection{Gaussian fitness noise} 

\begin{figure*}[t!]
\centering
\includegraphics[width = \hsize]{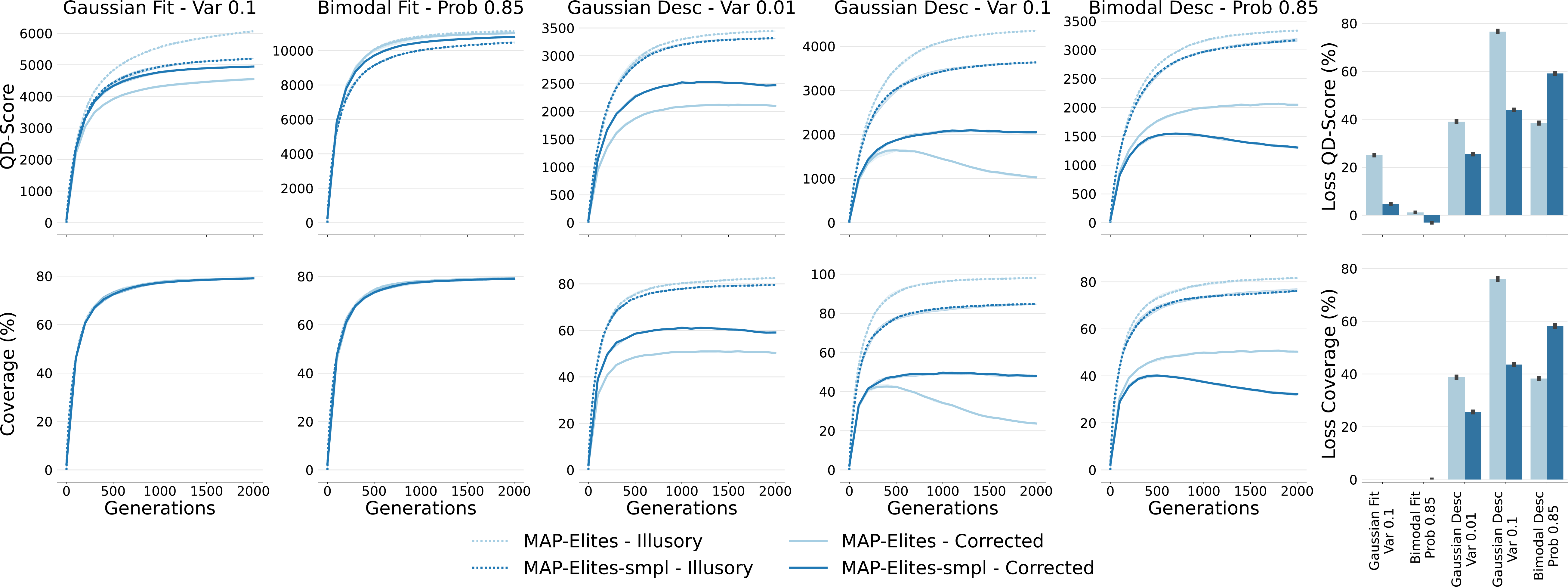}
\caption{
Results on the Performance-estimation benchmark tasks. Each column corresponds to one of the tasks from Table~\ref{tab:performance_task}. For each baseline, we give the Illusory (left - dashed line) and Corrected (left - plain line) QD-Score and Coverage \cite{uqd, aria}. Solid lines are the median across \replications{} replications, and shaded area the quartiles. We also give the Loss QD-Score (right - top) and Loss Coverage (right - bottom) from \citet{pga_stochastic}, that quantify the estimation capability of each baseline. 
}
\vspace{-1mm}
\label{fig:perf_results}
\end{figure*}

Gaussian noise on the fitness is the most commonly used noise distribution in previous \uqd{} works. 
Here, the variance parameter $\sigma^2$ for this distribution can take almost any value.
The results shows the benefit of sampling-based approaches in this case (Fig.~\ref{fig:perf_results}).

\subsection{Bimodal fitness noise} 

\label{sec:bimodalFitnessNoise}

One of the particularity of Gaussian distributions, as proposed in the previous section, is that sampling is always beneficial. 
Here, we propose a task where too low number of samples can actually be detrimental. We define a bimodal distribution with a distant low-probability second mode. This second mode is really detrimental for low number of samples, leading to an interesting phenomenon where MAP-Elites outperforms MAP-Elites-sampling (Fig.~\ref{fig:perf_results}).

\subsection{Small-Gaussian descriptor noise} 

Gaussian noise on the descriptor is the most commonly used noise distribution in previous \uqd{} works. The value of the variance $\sigma^2$ for this distribution is critical.
Here, we consider small values that lead individuals to stay with high probability within the same behavioural niche (i.e. MAP-Elites cell). 
The results show the benefit of sampling-based approaches in this case (Fig.~\ref{fig:perf_results}).

\subsection{Large-Gaussian descriptor noise} 

In this second version, we choose the standard deviation $\sigma$ to be sufficiently high to lead solutions to change behavioural niche with high probability.
This task makes the correct estimation of descriptor highly complex.
For instance, with a MAP-Elites grid of size $100\times 100$ and a value of $\sigma$ higher than $0.1$, most descriptor estimates with fewer than $100$ samples will not be placed in the appropriate cell.
The results highlight the limitation of MAP-Elites and MAP-Elites-sampling in this case (Fig.~\ref{fig:perf_results} and \ref{fig:performance_estimation_archives}).

\subsection{Bimodal descriptor noise} 

\begin{figure}[t!]
\centering
\includegraphics[width = \hsize]{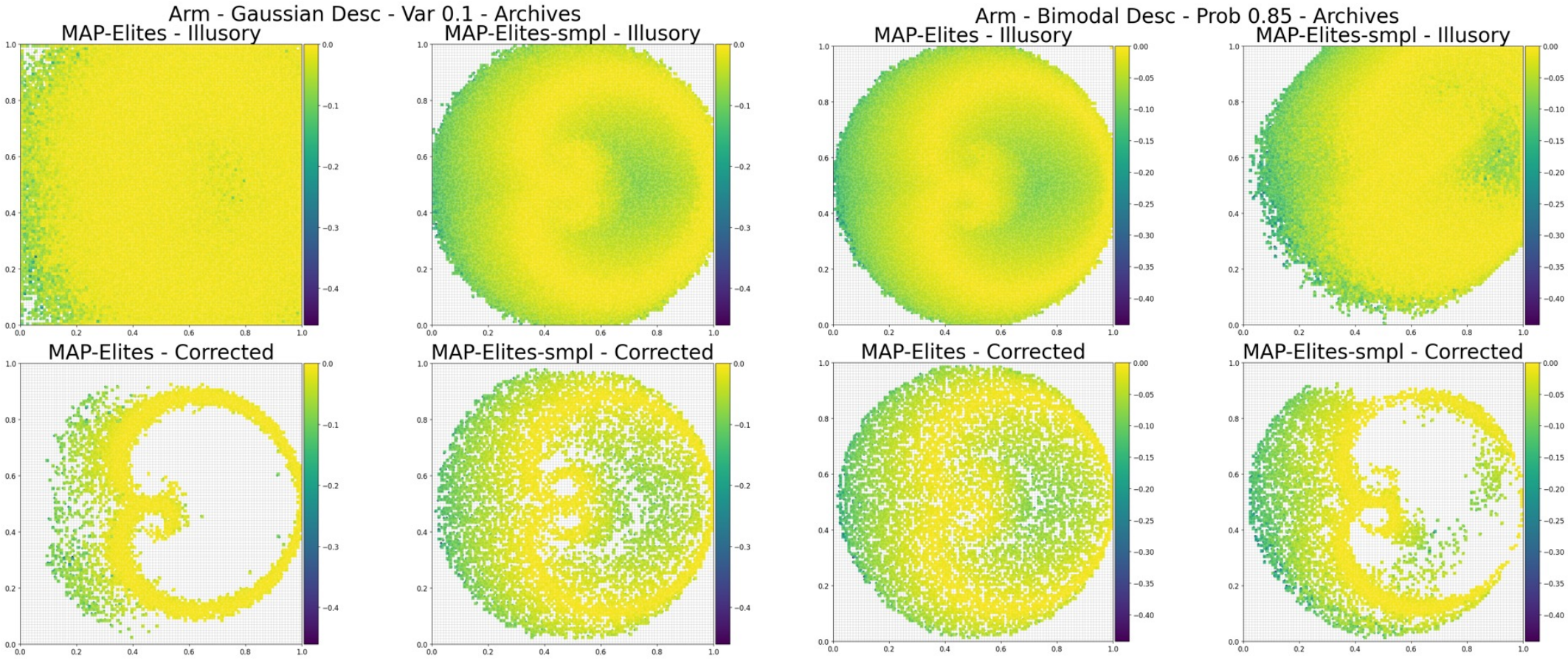}
\caption{
Illusory (top) and Corrected (bottom) archives for the two Performance-estimation descriptor noise tasks.
}
\vspace{-4mm}
\label{fig:performance_estimation_archives}
\end{figure}

The Bimodal distribution is the equivalent of the Bimodal Fitness Noise from Section~\ref{sec:bimodalFitnessNoise} for the descriptor. 
In this task, the noise takes the form of a bimodal distribution in 2 dimensions with a distant low-probability second mode (see Fig.~\ref{fig:arm_distributions}).
This distant second mode makes it difficult to estimate expected descriptors based on a low amount of samples, which is why MAP-Elites-sampling achieves the worst results in this task (Fig.~\ref{fig:perf_results} and~\ref{fig:performance_estimation_archives}).



\section{Reproducibility maximisation tasks} \label{sec:reproducibility}

In this section, we propose $2$ benchmark tasks that assess the ability of UQD approaches to keep solutions with high reproducibility. 
To simplify the problem, we consider only the reproducibility in term of descriptor, and we only focus on the Redundant Arm with zero-fitness from Section~\ref{sec:arm}. 
We introduce each task below, give corresponding equations in Table.~\ref{tab:reproducibility_task}, and some results in Figures~\ref{fig:reproducibility_results} and~\ref{fig:reproducibility_archives}. To quantify the performance of approaches, we display the Reproducibility-Score as defined by \citet{uqd}. Additionally, inspired by \citet{aria}, we add as baseline MAP-Elites-sampling-Reproducibility, which directly optimises for the reproducibility of solutions.

\subsection{Gaussian descriptor noise with $2$ possible variance values}

\begin{figure*}[t!]
\centering
\includegraphics[width = \hsize]{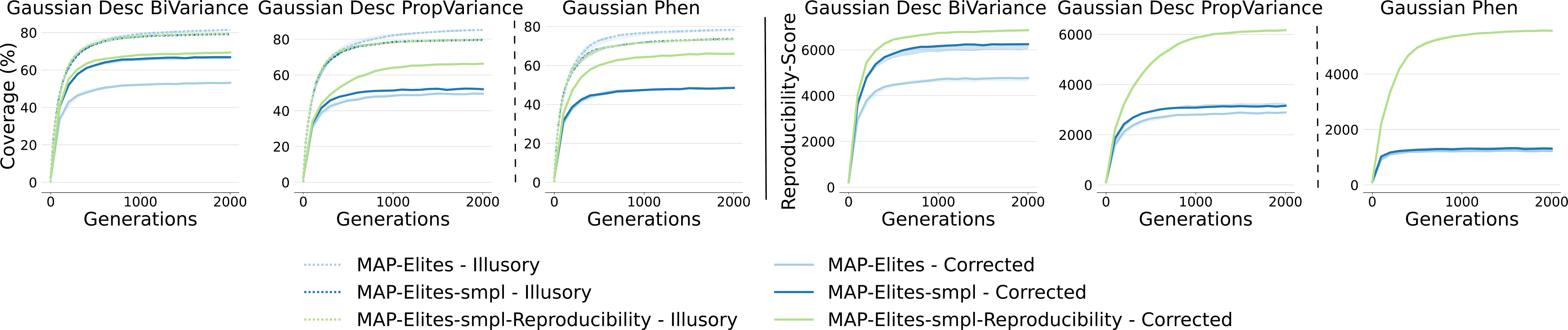}
\caption{
Results on the Reproducibility-maximisation and Realistic-combination benchmark tasks. Each column corresponds to one of the task defined in Table~\ref{tab:reproducibility_task} or~\ref{tab:realistic_task}. 
For each baseline, we give the Illusory Coverage (dashed line - left) and Corrected Coverage (plain line - left). We also display the Reproducible-Score from~\citet{uqd} that quantifies the ability of baselines to maintain reproducible solutions. It is computed based on the normalised variance $\sigma^2_{norm}$ as the sum over the Corrected archive of $1-\sigma^2_{norm}$.
Solid lines are the median across \replications{} replications, and shaded area the quartiles. 
}
\vspace{-1mm}
\label{fig:reproducibility_results}
\end{figure*}

\begin{figure}[t!]
\centering
\includegraphics[width = \hsize]{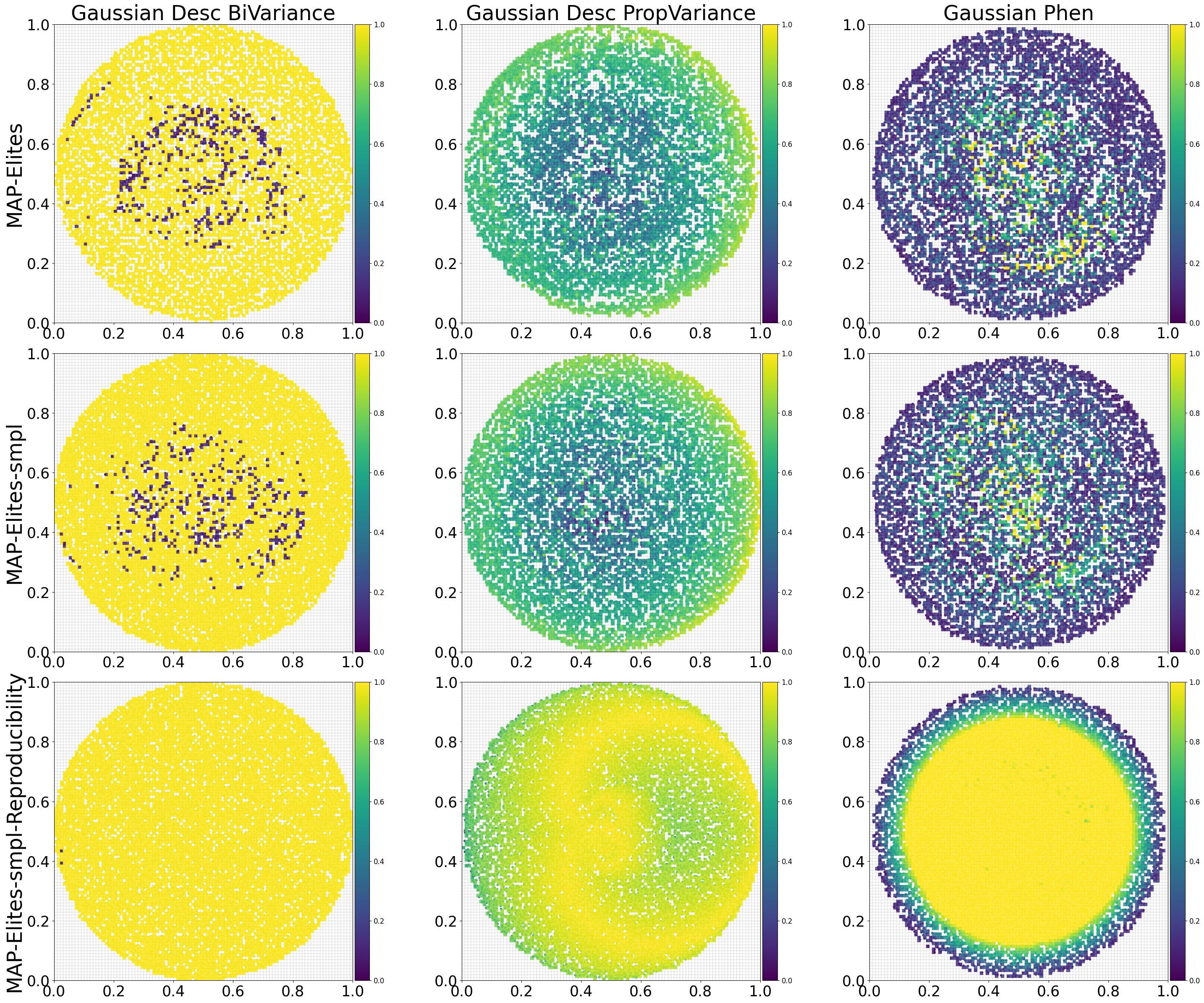}
\caption{
Reproducibility archives~\cite{uqd} for benchmark tasks of Reproducibility-maximisation and Realistic-combination. 
}
\vspace{-4mm}
\label{fig:reproducibility_archives}
\end{figure}

We first consider a simple task where there are two distinct possible values of variance for the descriptor: one small $\sigma_1^2$ and one large $\sigma_2^2$. 
Thus, solutions ending in the same cell can be either highly reproducible or poorly reproducible.
In this task, we attribute variance based on phenotype: we give low variance $\sigma_1^2$ to solutions with positive product of joint angles, and high variance $\sigma_2^2$ to the others.
Ideally, as the fitness is the same (zero) for all solutions, \uqd{} approaches are expected to prefer low-variance solutions to high-variance ones.
However, the results show that none of the two considered approaches manages to select reproducible solutions across the descriptor space, unlike MAP-Elites-sampling-Reproducibility \cite{aria} (Fig.~\ref{fig:reproducibility_results} and \ref{fig:reproducibility_archives}). 
\uqd{} algorithms require more complex mechanisms to be able to effectively solve this task \cite{adaptive, uqd, aria}.


\subsection{Gaussian descriptor noise with continuous variance values}

We propose a more complex version of the previous task, with the aim to refine the study of \uqd{} algorithms in reproducibility-maximisation tasks. 
In this task, the variance can take any values, continuously, within a range. Here, we attribute variance proportionally to the variance of the joint angles. 
Ideally, as the fitness is the same (zero) for all solutions, \uqd{} approaches are expected to select the most reproducible solutions for each cell. 
However, as for the previous task, results in Figure~\ref{fig:reproducibility_results} show that neither MAP-Elites nor MAP-Elites-sampling favour reproducible solutions. 

\section{Realistic combination task} \label{sec:realistic}

In addition to the targeted tasks proposed in the previous sections, we propose here one more realistic task that combines both the issues of performance estimation and reproducibility maximisation: the Gaussian Phenotype $J$ noise. We give the corresponding equations in Table ~\ref{tab:realistic_task} and some results in Figures~\ref{fig:reproducibility_results} and \ref{fig:reproducibility_archives}.

\subsection{Gaussian Phenotype $J$ noise}

In this task, we propose to change the \location{} to consider the phenotype (i.e. the joint angles).
Noise is applied on only one single pre-defined joint angle, and all solutions have zero-fitness.
UQD algorithms are expected to maximise the reproducibility of the solutions while also providing accurate estimates of their expected descriptors.
In this task, the best solutions are fully reproducible, i.e. have a descriptor variance of 0, as depicted on Figure~\ref{fig:arm_noise_on_6th_joint}. The results show that MAP-Elites and MAP-Elites-sampling do not manage to find these solutions (Fig.~\ref{fig:reproducibility_results} and \ref{fig:reproducibility_archives}).

\section{Conclusion and Discussion}

This paper presents the first set of \uqdlong{} benchmark tasks. Each of these tasks aims to highlight one limitation of existing QD algorithms in uncertain domains and points toward interesting directions of improvement. 
They are all defined for the Redundant Arm, a commonly used QD environment. 
In total, we propose $8$ tasks that can be split into $3$ main categories: Performance-estimation, Reproducibility-maximisation and Realistic tasks. The first two correspond to the two main issues that arise when considering Uncertain domains, as highlighted in previous work.
We hope that this paper will constitute a meaningful first benchmark for later advances in \uqd{} and that it will lead to more similar tasks to assess QD algorithms in Uncertain domains.

\bibliographystyle{ACM-Reference-Format}
\bibliography{references}

\end{document}